\documentclass[
]{ceurart}

\usepackage{latexsym}
\usepackage{url}
\usepackage{amsmath}
\usepackage{mathtools}
\usepackage{mdframed}
\usepackage{graphicx}
\usepackage{framed}
\usepackage{subcaption}
\usepackage{xcolor}

\newcommand\numberthis{\addtocounter{equation}{1}\tag{\theequation}}

\newcommand{\newchangecolor}{black}

\begin{document}

\copyrightyear{2021}
\copyrightclause{Copyright for this paper by its authors.
  Use permitted under Creative Commons License Attribution 4.0
  International (CC BY 4.0).}

\conference{ROMCIR 2021: Workshop on Reducing Online Misinformation through Credible Information Retrieval, held as part of ECIR 2021: the 43rd European Conference on Information Retrieval, March 28 – April 1, 2021, Lucca, Italy (Online Event)}

\title{Improving Authorship Verification using Linguistic Divergence}

\author[1]{Yifan Zhang}[%
email=yzhang114@uh.edu,
]

\author[1]{Dainis Boumber}[%
email=dainis.boumber@gmail.com,
]

\author[1]{Marjan Hosseinia}[%
email=ma.hosseinia@gmail.com,
]

\author[1]{Fan Yang}[%
email=fyang11@uh.edu,
]

\author[1]{Arjun Mukherjee}[%
email=arjun@cs.uh.edu,
]
\address[1]{University of Houston}

\begin{abstract}
We propose an unsupervised solution to the Authorship Verification task that utilizes pre-trained deep language models to compute a new metric called \emph{DV-Distance}. The proposed metric is a measure of the difference between the two authors comparing against pre-trained language models. Our design addresses the problem of non-comparability in authorship verification, frequently encountered in small or cross-domain corpora. To the best of our knowledge, this paper is the first one to introduce a method designed with non-comparability in mind from the ground up, rather than indirectly. It is also one of the first to use Deep Language Models in this setting. The approach is intuitive, and it is easy to understand and interpret through visualization. Experiments on four datasets show our methods matching or surpassing current state-of-the-art and strong baselines in most tasks.
\end{abstract}


\maketitle

\section{Introduction}
Authorship Attribution (AA) \cite{Stamatatos2009} and \normalfont{Verification} (AV) \cite{Luyckx2008} are challenging problems important in this age of "Fake News". The former attempts to answer \textit{who} wrote a specific document; the latter concerns itself with the problem of finding out whether the same person authored several documents or not. Ultimately, the goal of AV is to determine whether the same author wrote any two documents of arbitrary authorship. These problems have attracted renewed attention as we urgently need better tools to combat content farming, social bots and other forms of communication pollutions.

An interesting aspect of authorship problems is that technology used elsewhere in NLP has not yet penetrated it. Up until the very recent PAN 2018 and PAN 2020 Authorship event \cite{Kestemont, Bevendorff2020}, the most popular and effective approaches still largely relies on n-gram features and traditional machine learning classifiers, such as support vector machines (SVM) \cite{cortes1995support} and trees \cite{frery2014ujm}. Elsewhere, these methods recently had to give up much of their spotlight to deep neural networks. \textcolor{\newchangecolor}{ This phenomenon may be mostly attributed to the fact that authorship problems are often data constrained --- as the amount of text from a particular author is often very limited. From what we know, only a few deep learning models have been proposed and shown to be effective in authorship tasks \cite{Bagnall2015, DBLP:journals/corr/abs-1803-06456, BOUMBER18.535}, and even these networks require a good amount of text to perform well.} Likewise, transfer learning may not have been utilized to its full potential, as some of the recent work in deep language models shows it to be a silver bullet for tasks lacking training data \cite{howard2018universal}. 

We propose a deep authorship verification method that uses a new measurement, \emph{DV-Distance}. It estimates the magnitude and the direction of deviation of a document from the \emph{Normal Writing Style} (NWS) by modeling it with state-of-the-art language models such as the AWD-LSTM and RoBERTa architecture introduced in \cite{DBLP:journals/corr/abs-1708-02182, Liu2019}. We proposed an unsupervised method which directly utilize the DV-Distance and an supervised neural architecture which projecting these vectors into a separate space. These proposed models have an intuitive and theoretically sound architecture and comes with good interpretability. Experiments conducted on four PAN Authorship Verification datasets show our method surpass state-of-the-art in three and competitive in one.

\section{Authorship Verification and Non-comparability Problem}
\label{sec:problem}

In the following sections, we use the symbol $P$ to denote an authorship verification problem. Each problem $P$ consists of two elements: a set of \emph{known documents} $K$, and \emph{unknown documents}, $U$. Similarly, $k$ and $u$ represent a single known and unknown document, respectively. The task is then to find a hypothesis, $h$, that takes in both components and correctly estimates the probability that the same author writes them.
Important in many forensic, academic, and other scenarios, AV tasks remain very challenging due to several reasons. For one, in a cross-domain authorship verification problem, the documents in $K$ and $u$ could be of entirely different genre and type. More specifically, $K$ could contain several novels written by a known author, while $u$ is a twitter post. Another example demonstrating why a cross-domain model may be necessary is the case of a death note \cite{Stamatatos2015}, as it is implausible to obtain a set of $K$ containing death notes written by the suspect. Furthermore, solving an authorship verification problem usually involves addressing one or more types of limited training data challenges: a limited amount of training problems $P$, out-of-set documents and authors appearing in test data, or a limited amount of content in the document sets ${\{K,U\}}$ of a particular problem $P$. Many methods use sophisticated forms of test-time processing, data augmentation, or ensembling to successfully minimize these challenges' impact and achieve state-of-the-art results \cite{Bagnall2015, EasyChair:865}. However, such solutions typically result in prohibitively slow performance, most require a considerable amount of tuning, and almost all of them, to the best of our knowledge, require labeled data. As a result, existing methods are not relevant in many real-world scenarios.

\begin{figure}[h!]
\begin{mdframed}
\textbf{k:} I suppose that was the reason. We were waiting for you without knowing it. Hallo! \\
\textbf{u:} He maketh me to lie down in green pastures; he leadeth me beside the still waters.
\end{mdframed}
\caption{Sample document fragment from PAN-2015}
\label{sample doc 1}
\end{figure}

Based on our observations, it is not unusual for an authorship verification model to identify some salient features in either $K$ or $U$, yet fail to find a directly comparable case in the other member of the pair. An example consisting of two brief segments from different authors is shown in Figure \ref{sample doc 1}. We can immediately notice that document $u$ contains unusual words ``\emph{maketh}'' and ``\emph{leadeth}'' which are Old English. In contrast, document $k$ is written in relatively colloquial and modern English. A naive method of authorship verification one may devise in this scenario is to detect whether document $K$ contains the usage of ``\emph{makes}'', the modern counterpart to ``\emph{maketh}''. If there are occurrences of ``\emph{makes}'' in $K$, we may be able to conclude that the two documents are from different authors. The issue with this approach however, is the non-zero probability of $K$ containing no usages of ``\emph{makes}'' at all.

Although it is possible to overcome the problem of non-comparability hand-crafted features, feature engineering is often a labor-intensive process that requires manual labeling. It is also improbable to design all possible features that encode all characteristics of all words. On the other hand, while some modern neural network based methods built upon the concept of distributed representations (word embeddings), and was able to encode some of the essential features, there is no existing approach explicitly attempt to address the non-comparability problem.

To address the non-compatibility, we formulate \textit{Normal Writing Style} (NWS), which can be seen as a universal way to distinguish between a pair of documents and solve the AV task in most scenarios in an unsupervised manner. The documents difference or similarity is determined with respect to NWS; to this end, we establish a new metric called Deviation Vector Distance (DV-Distance). To the best of our knowledge, the proposed approach is the first model designed with non-compatibility in mind from the ground up.

\section{Normal Writing Style and Deviation Vector}
\label{sec:nws}

To make a small and often cross-domain document pair comparable, we propose to compare both documents to the \emph{Normal Writing Style} instead of directly comparing the pair. \textcolor{\newchangecolor}{ We can define the Normal Writing Style or NWS, loosely as what average writers would write on average, given a specific writing genre, era, and language. From a statistical perspective, the NWS can be modeled as the averaged probability distribution of vocabulary at a location, given its context. } As manifested in Figure \ref{sample doc 1}, the reason words \emph{maketh} and \emph{leadth} stand out in the documents $u$ is because they are rarely used in today's writing. They are hence \emph{deviant} from the Normal Writing Style.

\textcolor{\newchangecolor}{ We hypothesize that we can utilize modern neural language models to model NWS, and the predicted word embedding at a given location is a good semantic proxy of what an average writer would write at that location. And we also hypothesize that, generally, an author has a consistent direction of deviance in the word embedding space. } Consequently, if two documents $k$ and $u$ have the same direction of deviation, then the two documents are likely from the same author. Conversely, if two documents have a significantly different direction of deviation, then they are probably from different authors. Previous empirical evidence shows that word embedding constructed using neural language models are good at capturing syntactic and semantic regularities in language \cite{DBLP:journals/corr/abs-1301-3781, mikolov2013distributed, pennington2014glove}. The vector offsets encode properties of words and relationships between them. A famous example demonstrating these properties is the embedding vector operation: ``King - Man + Woman = Queen'', which indicates that there is a specific vector offset that encodes the difference in gender.

\begin{figure*}[]
\centering
\includegraphics[width=0.7\textwidth]{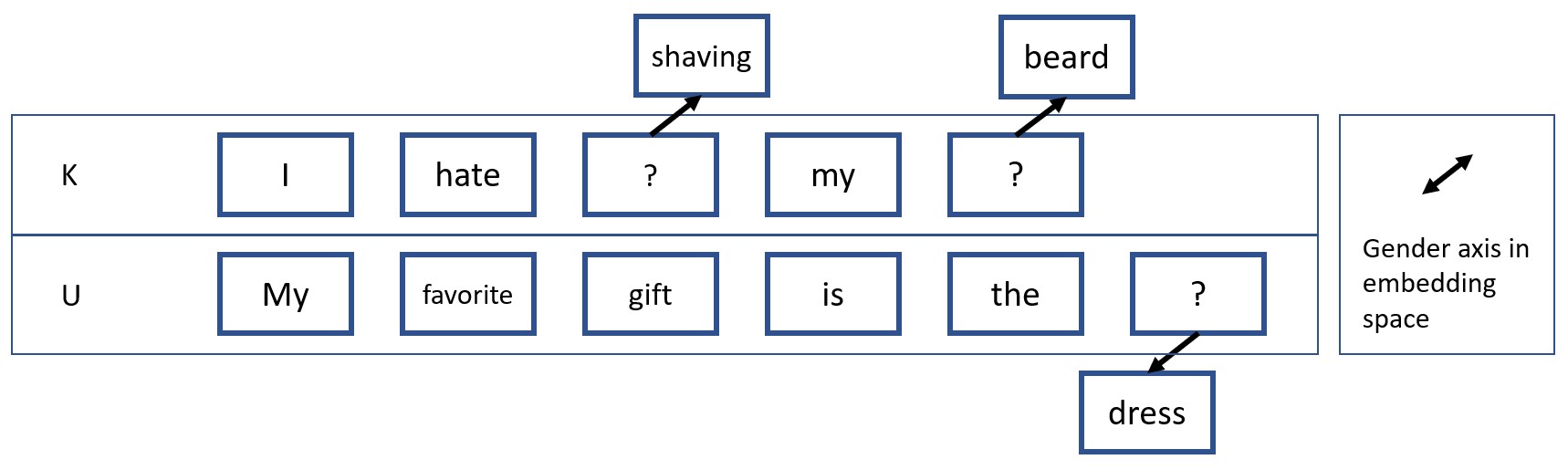}
\caption{Sample document fragments from PAN-2015}
\label{deviation direction}
\end{figure*}

Given the above context, we theorize it is possible to encode the deviance of \emph{maketh} from \emph{makes} as ``Maketh - Makes'' in a similar manner. We shall refer to the offset vector calculated this way as the Deviation Vector (DV). Figure \ref{deviation direction} shows an illustrative example that visualizes the roles of Normal Writing Style modeling and the DVs. In the upper part of the figure, a document $k$ by a male author is suggested, containing a sentence, "I hate shaving my beard." At the bottom half of the figure, we can see a document $u$ written by a female author: "My favorite gift is a dress." Assuming we have a NWS model that is able to correctly predict all the words except at locations marked using a question mark. In place of those words, NWS may predict very general terms, such as ``do'' or ``thing''. The actual words at these locations deviate from these general terms in the direction of the DV, represented in the figure using arrows. This specific example contains the words ``beard'' and ``dress'', usually associated with a particular gender, while the general terms are gender-less. The DV then must have a component along the direction of the gender axis in embedding space but in the opposite direction. 

\section{Language Model and Implementation Details}
\label{sec:model}
We used the AWD-LSTM architecture \cite{DBLP:journals/corr/abs-1708-02182}, implemented as part of Universal Language Model (ULMFit) \cite{howard2018universal}, and RoBERTa \cite{Liu2019} to model the Normal Writing Style. AWD-LSTM is a three-layered LSTM-based language model that trained by predicting the next word given the preceding sequence. Meanwhile, RoBERTa is a BERT-based model trained by predicting the masked word given an input sequence. Both of these language models are pre-trained on large corpuses and thus their predicted embedding for the unseen words can be used as a proxy of statistical distribution of \emph{Normal Writing Style}.

Assuming these language models can adequately model the \emph{Normal Writing Style}, the Deviation Vectors can be calculated by subtracting the actual embeddings of the words from the predicted word embeddings. More formally, for an input sequence consist of $n$ tokens $\{w_1, ..., w_n\}$. We use $EMB$ to denote the embedding layer of the language models, and use $LM$ to denote the language model itself. Then $EMB(w_i)$ and $LM(w_i)$ will correspond to the embedding of the actual token at location $i$ and the predicted embedding by the language model at location $i$ when the corresponding token is the next token (AWD-LSTM) or is masked (RoBERTa). The DV at location $i$ can then be calculated as:
\begin{equation}
    {DV_i = LM(w_i) - EMB(w_i)}
\end{equation}

\begin{figure}[t]
\includegraphics[width=\textwidth]{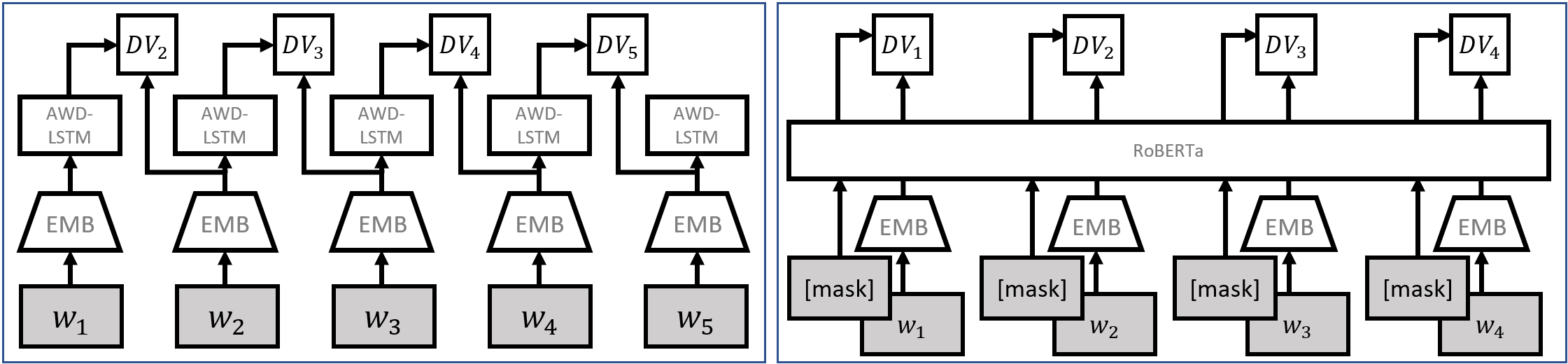}
\caption{A demonstration of the process of calculating DV using AWD-LSTM (left) and RoBERTa (right)}
\label{model-dv}
\end{figure}

Figure \ref{model-dv} demonstrates the respective processes of calculating the DVs for a given input sequence using AWD-LSTM and RoBERTa. For AWD-LSTM, at each token location $i$, the deviation vector is calculated by subtracting the predicted embedding generated at previous token location $i-1$, by the embedding of the current word at $i$. Consequently, for a document of $n$ words, a total of $n-1$ DVs can be generated. For RoBERTa, the predicted embedding at location $i$ is obtained by feeding the model complete sequence of text with the token at $i$ replaced by the ``[mask]'' token. A total of $n$ such inference need to be conducted to obtain all the predicted embeddings at each location. The DVs can then be calculated by subtracting the predicted embeddings using the actual token embeddings, resulting in a total of $n$ DVs.

\subsection{Unsupervised Method: DV-Distance}

To compare the direction of a deviation between two documents, we calculate the element-wise mean of all the DVs throughout each document to obtain the ``Averaged DVs''. For a given document of $n$ tokens, $ ADV(doc) = { \Sigma_{i=1}^{n} DV_i } / n $. Notice that for locations with a deviance between $LM$ and $EMB$, the corresponding $DV$ shall exert a larger influence on the document level $ADV$. Averaged DVs are calculated for both $K$ and $U$, then the DV-Distance can be calculated as the cosine similarity between $ADV(K)$ and $ADV(U)$.
\begin{equation}
    {DVDist(K, U) = \frac{ ADV(K) \cdot ADV(U) } { \left\Vert ADV(K) \right\Vert \left\Vert ADV(U) \right\Vert } }
\end{equation}

\textcolor{\newchangecolor}{ Since the DV-Distance method is completely unsupervised, the resulting distance values are relative instead of absolute. I.E., it is difficult to determine the classification result of a single document pair. Instead, a threshold value needs to be determined such that we can then classify all the document pairs with DV-Distance values greater than the threshold as "Not same author" and vice versa. To determine the threshold, we follow previous PAN winners such as \cite{Bagnall2015} and use the median of DV-distance value between all $K$ and $u$ pairs within the dataset as the threshold. Using this scheme is reasonable because PAN authorship verification datasets are guaranteed to be balanced. }
\textcolor{\newchangecolor}{ During our experiments, we found that the threshold value is relatively stable for a particular model in a given dataset, but can be quite different between LSTM and Bert-based models. For real-world applications, the threshold value can be determined ahead of time using a large dataset of similar genre and format as the problem to be evaluated. }

\subsection{Supervised Method: DV-Projection}
One of the major deficiencies of our Deviation Vector theory is that it assumes all differences in the DV hyperspace are relevant. However, one can imagine this assumption does not always hold in all the authorship verification settings. For example, the gender dimension shift shown in Figure \ref{deviation direction} can be a useful clue when conducting authorship verification on a Twitter dataset or in the context of autobiographies. It may be less relevant if the gender shift occurs in a novel, as the vocabularies used in the novel are more relevant to its characters' genders instead of the author's.

To address this issue, we propose to use a supervised neural network architecture to project the DVs onto axes that are most helpful for distinguishing authorship features. As we will demonstrate in the results and analysis section of this work, these DV projections are very  effective when combining with the original token embeddings generated using the language models.

\begin{figure}[t]
\centering
\includegraphics[width=0.7\textwidth]{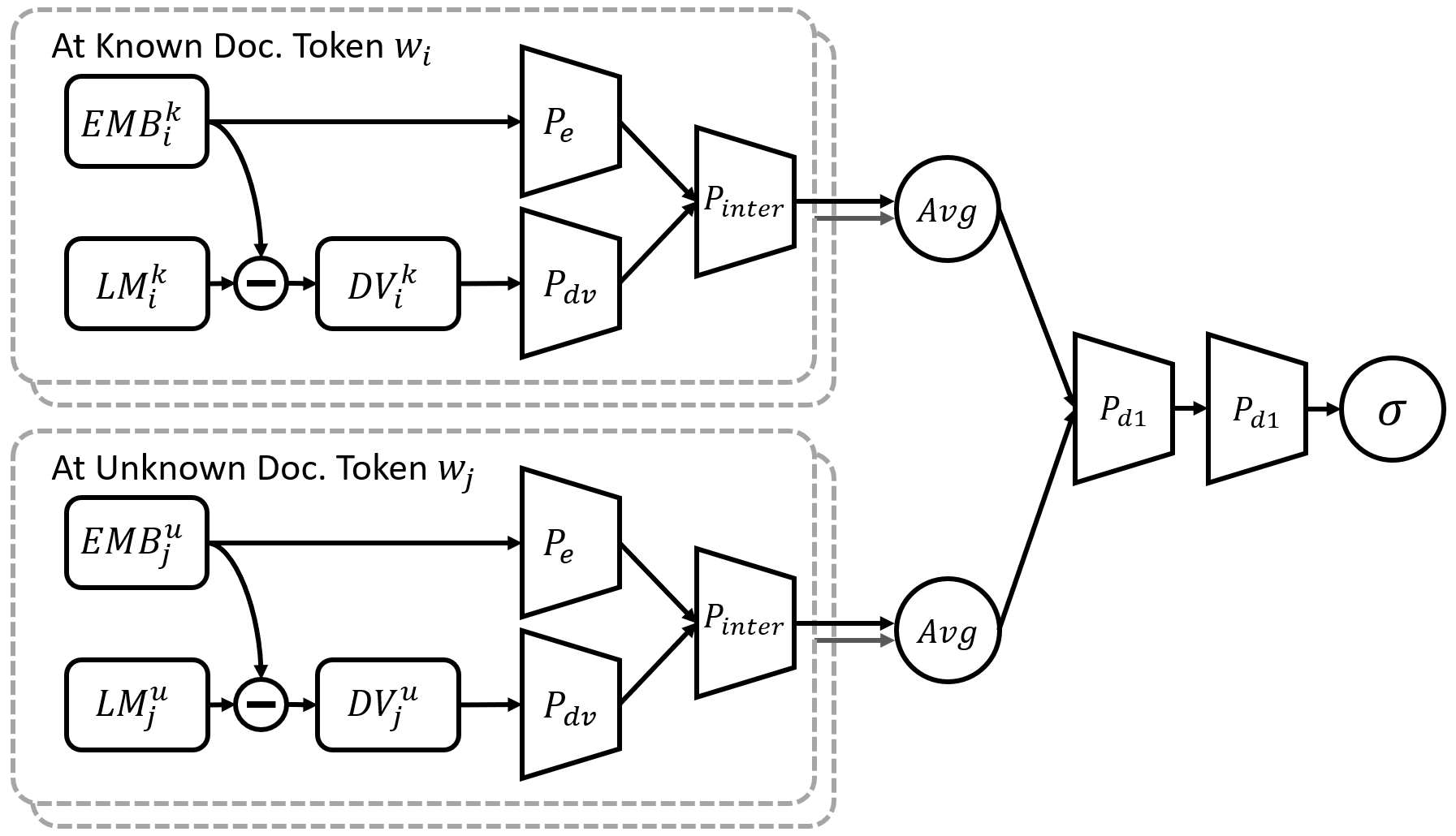}
\caption{ \textcolor{\newchangecolor}{ Network architecture of the DV-Projection method. Vectors $EMB$, $LM$ and $DV$ are represented using rounded rectangle shape. Fully connected layers are represented using trapezoid shape. Element-wise math operations are represented using circles.} }
\label{model-dvproj}
\end{figure}

Here we shall formally define the DV-Projection process. Given we have the embeddings and DVs for both a known document and an unknown document, each denoted using $EMB_i^k$, $DV_i^k$, $EMB_i^u$, $DV_i^u$. We use dense layers $P_e$ and $P_{dv}$ with embeddings and DVs respectively to extract prominent features. These features are then feed together into dense layer $P_{inter}$. The outputs of $P_{inter}$ are then average-pooled along the sequence to produce document-level features. Lastly, features from both known and unknown documents are connected to 2 additional fully-connected layers $P_{d1}$, $P_{d2}$ to produce the final output. These operations can be summarized in equation \ref{f:supervised} and visualized in figure \ref{model-dvproj}, all layers are used in combination with hyperbolic tangent as activation function:

\begin{align*}
 \label{f:supervised}
 TokenFeature_i^k &= P_{inter}( P_e( EMB_i^k ), P_{dv}( DV_i^k ) ) \numberthis \\
 TokenFeature_j^u &= P_{inter}( P_e( EMB_j^u ), P_{dv}( DV_j^u ) ) \\
 DocFeature^k &= AvgPool( TokenFeature^k ) \\
 DocFeature^u &= AvgPool( TokenFeature^u ) \\
 logit &= P_{d2}( P_{d1}( DocFeature^k, DocFeature^u ) )
\end{align*}

\textcolor{\newchangecolor}{ To allow training of the above model together with RoBERTa, we breaks documents from the original training document pairs into segments of 128 tokens long. We then build smaller training example pairs from these short document segments and label them accordingly. This approach not only allows us to build a lot more training examples to properly train the network parameters, it also forces the model to be more robust by limiting the amount of text it has access to. The training loss used is binary cross entropy loss in combination with the Sigmoid function. }

\textcolor{\newchangecolor}{ Because the DV-Projection method is a supervised model, from a theoretical perspective the model can learn the optimal threshold for classification, therefore eliminating the needs for using median value as threshold. However, the document segment based training pair generation method can generates significantly more ``same author'' pairs than ``different author'' pairs. Therefore the resulting trained model is biased and cannot be assumed to have a 0 valued threshold\footnote{ \textcolor{\newchangecolor}{ In real-world application this problem can be easily addressed by simply generating a large and balanced training dataset. } }. To make it consistent, we also use the testing set median value as the threshold for DV-Projection method\footnote{ \textcolor{\newchangecolor}{ One can also opt to use training set median value as the threshold. To give an rough impression of how this will impact the performance: On PAN14N dataset, using testing set median value as threshold will produce 61\% in accuracy, using training set median value as threshold will produce 65\% in accuracy. On PAN14E dataset: using testing set median value as threshold will produce 73\% in accuracy, using training set median value as threshold will produce 70\% in accuracy. } }. }

\section{Experiments}

\label{sec:experiments}

The goal of the empirical study described in the following section is to validate the proposed DV-Distance and DV-Projection method. For this purpose, we use authorship verification datasets released by PAN in 2013 \cite{juola2013overview}, 2014 \cite{stamatatos2014overview} and 2015 \cite{stamatatos2015overview}.

\subsection{Datasets}

The 2013 version of PAN dataset consists of 10 training problems and 30 testing problems. PAN 2014 includes two separate datasets, Novels and Essays. PAN 2014N consists of 100 English novel problems for training and 200 English problems for testing. PAN 2014E consists of 200 English essay problems for training and 200 English essay problems for testing. PAN 2015 is a cross-topic, cross-genre author verification dataset, which means known documents and an unknown document may come from different domains. PAN 2015 contains 100 training problems and 500 testing problems.\\

\subsection{Evaluation Metrics}
For each PAN dataset, we follow that year's challenge rules. PAN 2013 uses accuracy, Receiver-Operating Characteristic (ROC) and $Score = Accuracy \times ROC$. PAN 2014 introduces the c@1 measure to replace accuracy to potentially reward those contestants who choose not to provide an answer in some circumstances. This metric was proposed in \cite{Penas2011}, and it is defined as
\begin{equation} \label{catone}
{c@1 = (\frac{1}{n}) \times (n_c + ({n_u \times \frac{n_c}{n}}))},
\end{equation} 

where $n_c$ is the number of problems correctly classified, and $n_u$ is the number of open problems. The Score for PAN 2014 and 2015 is calculated as the product of c@1 and ROC, $c@1 \times ROC$.

\label{sec:analysis}

\begin{table*}[ht!]
\centering

\scalebox{0.8}{
\begin{tabular}{ ll | lll | lll }
  \hline
   &  & \multicolumn{3}{c|}{PAN14E} & \multicolumn{3}{c}{PAN14N}\\
   Category & Method & c@1 & ROC & Score & c@1 & ROC & Score \\
  \hline
  Baseline & GNB & 0.675 & 0.741 & 0.5   & 0.56  & 0.743 & 0.416 \\
  Baseline & LR  & 0.675 & 0.728 & 0.491 & 0.515 & 0.604 & 0.311 \\
  Baseline & MLP & 0.7 & 0.768 & 0.538 & 0.54 & 0.782 & 0.422 \\
  PAN & FCMC \cite{Modaresi2014} & 0.58 & 0.602 & 0.349 & 0.71 & 0.711 & 0.508 \\
  PAN & Frery \cite{Frery2014} & 0.71 & 0.723 & 0.513 & 0.59 & 0.61  & 0.36  \\
      & TE \cite{DBLP:journals/corr/abs-1803-06456} & 0.67 & 0.675 & 0.452 & 0.695& 0.7   & 0.487 \\
      & 2WD-UAV \cite{EasyChair:865} & \textbf{0.73} & 0.761 & 0.555 & 0.68 & \textbf{0.801} & 0.552 \\
  Our model & \textbf{DV-Dist. L} &  0.58 & 0.575 & 0.334 & \textbf{0.82} & 0.79  & \textbf{0.648}\\
  Our model & \textbf{DV-Dist. R} &  0.52 & 0.526 & 0.274 & 0.71 & 0.739  & 0.525 \\
  Our model & \textbf{DV-Proj. R} &  \textbf{0.73} & \textbf{0.778} & \textbf{0.569} & 0.61 & 0.668  & 0.41 \\
  \hline
   &  & \multicolumn{3}{c|}{PAN13} & \multicolumn{3}{c}{PAN15}\\
   Category & Method & Acc. & ROC & Score & c@1 & ROC & Score \\
  \hline
  Baseline & GNB & 0.633 & 0.795 & 0.503 & 0.552 & 0.78  & 0.431 \\
  Baseline & LR  & 0.7   & 0.781 & 0.547 & 0.544 & 0.796 & 0.433 \\
  Baseline & MLP & 0.533 & 0.5   & 0.267 & 0.554 & 0.687 & 0.381 \\
  PAN & MRNN \cite{Bagnall2015} &  -  &  -  &  -  & \textbf{0.76} & 0.81 & 0.61 \\
  PAN & Castro \cite{Castro2015} &  -  &  -  &  -  & 0.69 & 0.75 & 0.52 \\
  PAN & GenIM \cite{Seidman2013} & 0.8 & 0.792 & 0.633 &  -  &  -  &  -  \\
  PAN & CNG \cite{Jankowska} &  -  & \textbf{0.842}  &  -  &  -  &  -  &  -  \\
      & TE \cite{DBLP:journals/corr/abs-1803-06456} & 0.8 & 0.835 & 0.668 & 0.748 & 0.75  & 0.561  \\
      & 2WD-UAV \cite{EasyChair:865} & \textbf{0.82} & 0.825 & \textbf{0.677} & 0.75 & 0.822 & 0.617 \\
   Our model & \textbf{DV-Dis. L} & 0.7 & 0.763 & 0.534 & \textbf{0.76} & \textbf{0.834} & \textbf{0.634} \\
   Our model & \textbf{DV-Dis. R} & 0.63 & 0.746 & 0.472 & 0.716 & 0.767 & 0.548 \\
  \hline
\end{tabular}
}

\caption{ \textcolor{\newchangecolor} { Authorship Verification results for PAN datasets. } }
\label{ResultsTable}

\end{table*}

\subsection{Baselines}
\textbf{Classic Models with N-gram Features:} In our study we use a set of baselines reported in \cite{DBLP:journals/corr/abs-1803-06456}. They are produced using seven sets of features, including word n-grams, POS n-grams, and character 4-gram. The features need to be transformed because baselines are standard classification algorithms. According to the authors, simple concatenation of two documents' features produces poor results, and use seven different functions to measure the similarity between feature vectors from both documents, including \emph{Cosine Distance}, \emph{Euclidean Distance}, and \emph{Linear Kernel}. Several common classifiers are trained and evaluated using these similarity measurements, providing a reasonable representation of the performance that is achievable using classic machine learning models and n-gram feature sets. Out of all the baseline results, three classifiers with the highest performance are reported along with the other PAN results for comparison. The selected classifiers are Gaussian Naive Bayes (GNB), Logistic Regression (LR) and Multi-Layer Perceptron (MLP). We compare them with the proposed approach along with the state-of-the-art methods.

\textbf{PAN Winners:} We compare our results to the best performing methods submitted to PAN each year. The evaluation results of the participant teams are compiled in the overview reports of PAN 2013 \cite{Juola2013}, 2014 \cite{Stamatatos2014} and 2015 \cite{Stamatatos2015}. In \textbf{PAN 2013}, the best-performing methods are the General Imposters Method (GenIM) proposed by \cite{Seidman2013} and the Common N-Gram (CNG) dissimilarity measure proposed by \cite{Jankowska}. In \textbf{PAN 2014} challenge, the best method for English Essay dataset is proposed by \cite{Frery2014} (Frery), and the best method for English Novel dataset is by \cite{Modaresi2014} which uses Fuzzy C-Means Clustering (FCMC). In \textbf{PAN 2015}, the Multi-headed Recurrent Neural Networks (MRNN) proposed in \cite{Bagnall2015} outperforms the second best submission (Castro) \cite{Castro2015} of the same year by a large margin.

\textbf{Transformation~Encoder:} In \cite{DBLP:journals/corr/abs-1803-06456}, an auto-encoder based authorship verification model performed competitively on PAN. We include its results to evaluate our model against one of the newest and strongest performers.

\textbf{2WD-UAV:} A language modeling based approach that relies on transfer learning an ensemble of heavily regularized deep classification models and data augmentation shows state-of-the-art performance, surpassing all verification methods evaluated on PAN that we are aware of \cite{EasyChair:865}. Like our approach, it is based on a deep language model; however, it is otherwise similar to the majority of solid AV performers. 

\section{Results and Discussion}
\label{sec:results}

Table \ref{ResultsTable} shows the results from experiments on PAN datasets, detailed in Section \ref{sec:experiments}.
The proposed unsupervised DV-Distance method conducted using AWD-LSTM and RoBERTa is denoted as ``DV-Dist. L'' and ``DV-Dist. R'', respectively. The proposed supervised DV-Projection method is trained using DVs produced by RoBERTa and is labeled as ``DV-Proj. R'' in the table. We were only able to train the projection model on PAN14E and PAN14N due to both of them have relatively large training set.

\textcolor{\newchangecolor}{For PAN 2013, our results are slightly below the best performer of that year in terms of accuracy and AUC-ROC; the 0.1 difference in accuracy translates to 3 problems difference out of 30 testing problems. The PAN 2013 corpus are text segments from published Computer Science textbooks. The best performing model in this dataset is the neural network-based model from 2WD-UAV. }

\textcolor{\newchangecolor}{For PAN 2014, we observed some interesting results. For the Novels part of the challenge, our unsupervised DV-Distance method based on LSTMs drastically improves upon previous state-of-the-art models, surpasses the previous best result by 18 percent. On the other hand, for the Essay dataset, both unsupervised DV-Distance methods failed to capture the feature necessary to complete the task, showing only 58\% and 52\% in accuracy. However, the supervised DV-Projection method successfully projects the DVs generated using RoBERTa into a hyperspace that is suitable for the essay AV problems, resulting in significant performance improvement over the unsupervised models and slightly outperforms the previous best result from 2WD-UAV.}

\textcolor{\newchangecolor}{PAN 2015 edition places its focus on cross-genre and cross-topic authorship verification task. Based on our observations, the corpus mainly consists of snippets of novels of different genres and sometimes poems. Our proposed DV-Distance method based on multi-layer LSTMs once again shows excellent performance in this dataset, slightly outperforms the previous best model MRNN \cite{Bagnall2015}. In cross-domain settings like PAN 2015, the problem of non-comparability is likely to be very pronounced. The strong performance of our methods in this dataset therefore verifies that these methods are quite robust against domain shift and non-comparability.}

\textcolor{\newchangecolor}{Overall, we have observed two consistent trends in our experiments. First, we find that the AWD-LSTM based DV-Distance method consistently performs better than the RoBERTa based DV-Distance method. At first glance, this may seems counter-intuitive, as BERT-based models are generally regarded as one of the best performing model for language modeling. We theorize that this is precisely the culprit: RoBERTa was able to predict the target word much more accurately, both due to its architectural advantage and it simply has access to more contextual information. However, if the language model is performing ``too accurate'', it failed to act as a model which represents averaged writing style, but instead mimicking the author's tone and style. From a mathematical perspective, predictions that are ``too accurate'' will cause $DV$s calculated using equation (1) to have a magnitude close to zero, then later steps in equation (2) or (3) will have very little information to work with.}

\textcolor{\newchangecolor}{Second, we find that our proposed methods are most suitable for novel and fiction-type documents. Our methods demonstrated state-of-the-art performance in both PAN 2014 Novel and PAN 2015; both consist of mainly novel documents. On the other hand, PAN 2013 and PAN 2014 essay contains writing styles that are more formal and academic-oriented, for which our models performed less competitive. We theorize that this is because essay documents are easier to predict, whereas novels are much more ``unpredictable''. This difference in predictability means in novel datasets, we can obtain higher quality DVs; while in essay datasets, the language models are once again making predictions that are ``too accurate'', corroborating the first theory we discussed above.}

\begin{figure}[t]
\centering
  \begin{subfigure}[b]{.50\linewidth}
    \centering
    \includegraphics[width=\textwidth]{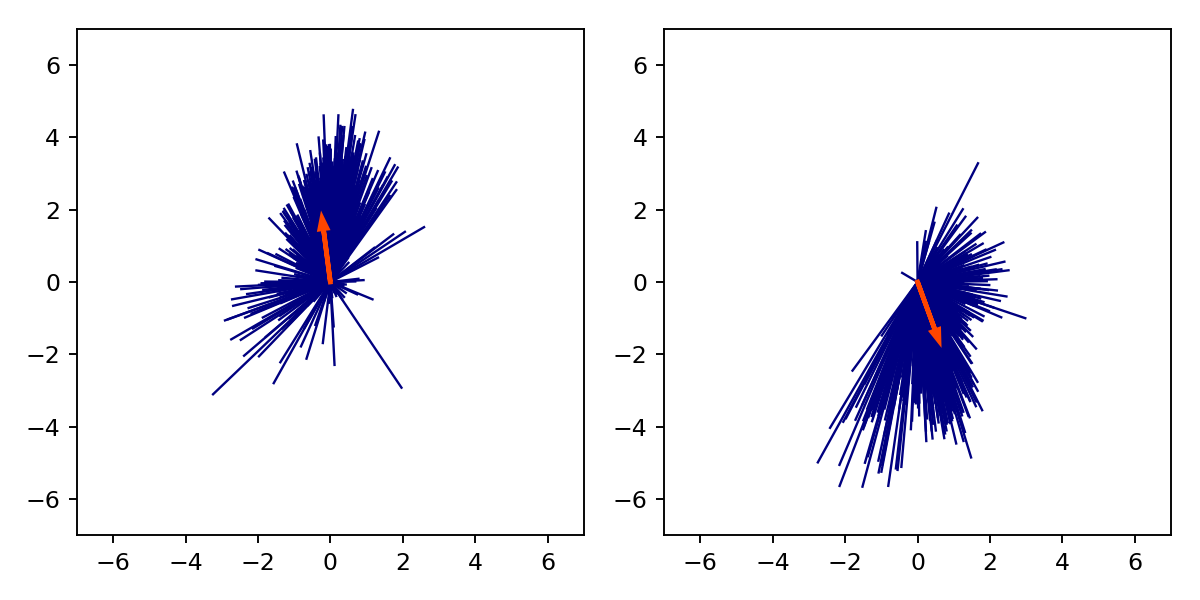}
    \caption{DVs of a document pair by different authors.}
    \label{fig:flowerplot_diff}
  \end{subfigure}\\%
  \begin{subfigure}[b]{.50\linewidth}
    \centering
    \includegraphics[width=\textwidth]{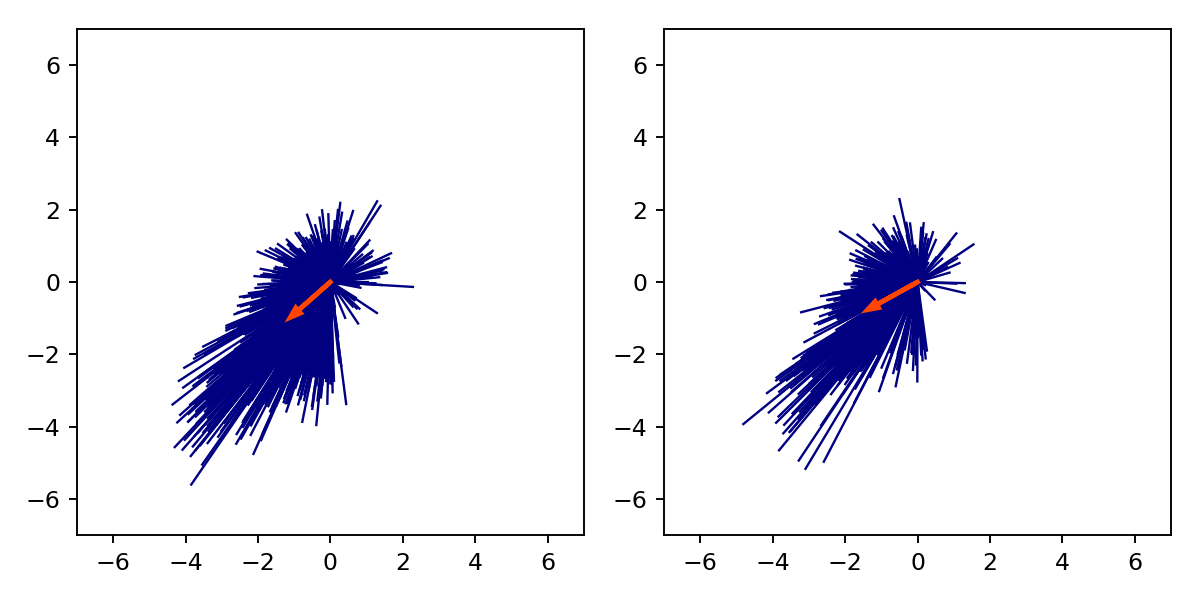}
    \caption{DVs of a document pair by the same author.}
    \label{fig:flowerplot_simi}
  \end{subfigure}
\caption{Visualization of deviation vectors in 2D. Each line corresponds to a word level DV and all words in a document is visualized in one subplot. The arrows in each subplot represents the averaged DV direction of that document.}
\label{fig:flowerplot}
\end{figure}

Deviation vectors of two PAN 2015 document pairs are visualized in Figure \ref{fig:flowerplot}. Figure \ref{fig:flowerplot_diff} shows two documents from different authors while Figure \ref{fig:flowerplot_simi} shows two documents by the same author. The plots are generated by conducting PCA on the DVs at each word, projecting the 400 dimension DVs from AWD-LSTM to 2 dimension. A longer line in the plots hence represents a bigger deviation from the NWS. We can observe that in Figure \ref{fig:flowerplot_diff} the DVs' directions are in opposite direction while in Figure \ref{fig:flowerplot_simi} their directions are similar.

\section{Related Work}

Much of the existing work in authorship verification is based on vocabulary distributions, such as n-gram frequency. The hypothesis behind these models is that the relative frequencies of words or word combinations can be used for profiling the author's writing style \cite{Stamatatos2009, Hoover2001}. One can conclude that two documents are more likely to be from the same author when the distributions of the vocabularies are similar. For example, in one document we may find that the author frequently uses ``\emph{I like ...}'', while in another document the author usually writes ``\emph{I enjoy ...}''. Such a difference may probably indicate that the documents are from different authors. This well-studied approach has had many successes, such as settling the dispute of "Federalist Papers" \cite{rudman2012twelve}. However, its results are often less than ideal when dealing with a limited data challenge.

The amount of documents in $K$ and $U$ is often insufficient to build two uni-gram word distributions that are comparable, let alone 3-gram or 4-gram ones. The depth of difference between two sets of documents is often measured using the unmasking technique while ignoring the negative examples \cite{koppel2004authorship}. This one-class technique achieves high accuracy for 21 considerably large (over 500K) eBooks. A simple feed-forward three layer auto-encoder (AE) can be used for AV, considering it a one-class classification problem \cite{manevitz2007one}. Authors observe the behavior of the AE for documents by different authors and build a classifier for each author. The idea originates from one of the first applications of auto-encoders for novelty detection in classification problems \cite{japkowicz1995novelty}. 

AV is studied for detecting linguistic traits of sock-puppets to verify the authorship of a pair of accounts in online discussion communities \cite{kumar2017army}. A spy induction method was proposed to leverage the test data during the training step under "out-of-training" setting, where the author in question is from a closed set of candidates while appearing unknown to the verifier \cite{Hosseinia2017DetectingSI}. 

In a more realistic setting, we have no specified writing samples of a questioned author, and there is no closed candidate set of authors. Since 2013, a surge of interest arose for this type of AV problem. \cite{seidman2013authorship} investigate whether one document is one of the outliers in a corpus by generalizing the Many-Candidate method by \cite{koppel2011authorship}. The best method of PAN 2014E optimizes a decision tree. Its method is enriched by adopting a variety of features and similarity measures \cite{frery2014ujm}. For PAN 2014N, the best results are achieved by using fuzzy C-Means clustering \cite{modaresi2014language}. In an alternative approach, \cite{koppel2014determining} generate a set of impostor documents and apply iterative feature randomization to compute the similarity distance between pairs of documents. One of the more exciting and powerful approaches investigates the language model of all authors using a shared recurrent layer and builds a classifier for each author \cite{bagnall2015author}. Parallel recurrent neural network and transformation auto-encoder approaches produce excellent results for a variety of AV problems \cite{DBLP:journals/corr/abs-1803-06456}, ranging from PAN to scientific publication's authorship attribution \cite{BOUMBER18.535}. In 2017, a non-Machine Learning model comprised of a compression algorithm, a dissimilarity method, and a threshold was proposed for AV tasks, achieving first place in two of four challenges \cite{halvani2017usefulness}.

\textcolor{\newchangecolor}{Among the models mentioned above, MRNN proposed in \cite{Bagnall2015} is the most comparable method to what we have introduced in this work. MRNN is an RNN-based character-level neural language model that models the flow of the known author documents $K$ and then is applied to the unknown document $u$. If the language model proves to be pretty good at predicting the next word on the unknown document (lower cross-entropy), then one can conclude they are likely written by the same author. While both MRNN and our DV-Distance-based methods utilize neural language modeling, for MRNN the language model represents a specific author's writing style and need to be trained on the corpus $K$. In practice, training a language model on a small corpus without overfitting can be very challenging, if not impossible. In contrast, the DV-Distance methods proposed in this work does not require training a author-specific language model, instead, both known and unknown documents are compared against a common language model, allowing for evaluation on AV problems with shorter documents.}

\section{Conclusion}
 
In this paper, we present a novel approach to the authorship verification problem. Our method relies on using deep neural language models to model the Normal Writing Style and then computes the proposed DV-Distance between the set of known documents and the unknown document. The evaluation shows that authorship style difference strongly correlated with the distance metric we proposed. Our method outperforms several state-of-the-art models on multiple datasets, both in terms of accuracy and speed.

\section{Acknowledgement}
\textcolor{\newchangecolor} {
Research was supported in part by grants NSF 1838147, NSF 1838145, ARO W911NF-20-1-0254. The views and conclusions contained in this document are those of the authors and not of the sponsors. The U.S. Government is authorized to reproduce and distribute reprints for Government purposes notwithstanding any copyright notation herein.
}
\bibliography{mybib}

\appendix

\end{document}